{

\documentclass[journal]{IEEEtran}

\usepackage[dvipsnames]{xcolor}
\usepackage{algorithm}
\usepackage{algpseudocode}
\usepackage{multirow}
\usepackage{amsmath}
\usepackage{mathtools}
\usepackage{graphicx}
\usepackage{xcolor}
\usepackage{enumitem,kantlipsum}

\usepackage[normalem]{ulem}
\setlist{nolistsep}

\ifCLASSINFOpdf
\else
\fi
\hyphenation{op-tical net-works semi-conduc-tor}

 \usepackage{amssymb}
\usepackage{pifont}
\usepackage{tikz}

\usepackage{float,booktabs,array}

\newcommand{\FIRSTEA}{\textcolor{black}{ES-p-MCTS}}
\newcommand{\SECONDEA}{\textcolor{black}{ES-MCTS}}

\begin{document}
%


\title{On the Evolution of the MCTS Upper Confidence Bounds for Trees by Means of Evolutionary Algorithms in the Game of Carcassonne}

%
%
%

\author{Edgar Galv\'an$^*${\thanks{$^*$Leading and corresponding author.}} and Gavin Simpson
\thanks{Edgar Galv\'an and Gavin Simpson are both with the Naturally Inspired Computation Research Group and with the Department
of Computer Science, Maynooth University, Lero, Ireland  e-mails: edgar.galvan@mu.ie and gavin.simpson.2021@mumail.ie}}

\markboth{}
\markboth{}

%



\maketitle

\begin{abstract}
 Monte Carlo Tree Search (MCTS) is a sampling best-first method to search for optimal decisions. The MCTS’ popularity is based on its extraordinary results in the challenging two-player based game Go, a game considered much harder than Chess and that until very recently was considered infeasible for Artificial Intelligence methods. The success of MCTS depends heavily on how the tree is built and the selection process plays a fundamental role in this. One particular selection mechanism that has proved to be reliable is based on the Upper Confidence Bounds for Trees, commonly referred as UCT. The UCT attempts to nicely balance exploration and exploitation by considering the values stored in the statistical tree of the MCTS. However, some tuning of the MCTS UCT is necessary for this to work well. In this work, we use Evolutionary Algorithms (EAs) to evolve mathematical expressions with the goal to substitute the UCT mathematical expression. We compare our proposed approach, called Evolution Strategy in MCTS (ES-MCTS) against five variants of the MCTS UCT, three variants of the *-minimax family of algorithms as well as a random controller in the Game of Carcassonne. We also use a variant of our proposed EA-based controller, dubbed ES partially integrated in MCTS. We show how the ES-MCTS controller, is able to outperform all these 10 intelligent controllers, including robust MCTS UCT controllers.

\end{abstract}


%
\IEEEpeerreviewmaketitle

\section{Introduction}

\label{sec:introduction}

Monte Carlo Tree Search (MCTS) is a sampling method for finding \textit{optimal decisions} by performing random samples in the decision space and building a tree according to partial results.  In a nutshell, Monte Carlo methods work by approximating future rewards that can be achieved through random samples. The evaluation function of MCTS relies directly on the outcomes of simulations. Thus, the accuracy of this function increases by adding more simulations. The optimal search tree is guaranteed to be found with infinite memory and computation~\cite{kocsis2006bandit}. However, in more realistic scenarios, MCTS can produce beneficial approximate solutions.

MCTS has gained popularity in two-player board games partly thanks to its success in the game of Go~\cite{alphago}, including beating professional human players. The diversification of MCTS in other research areas is extensive. For instance, MCTS has been explored in energy-based problems~\cite{Galvan_EnergyCon_2014,galvan2014heuristic} and in the design of deep neural network (DNN) architectures~\cite{wang2019alphax}. These two extreme examples demonstrate the successful versatility, use and applicability of MCTS in different problems.

The success or failure of MCTS depends heavily on how the MCTS statistical tree is built. The selection policy, responsible for this, behaves incredibly well when using the Upper Confidence Bounds for Trees~\cite{10.1007/11871842_29}, commonly referred as UCT. Some conditions are to be met for this to work well such as having enough number of simulations and tuning a value responsible for balancing exploration. In this work, we use EAs, in particular Evolution Strategies (ES)~\cite{Rechenberg10.1007/978-3-642-83814-9_6} to evolve mathematical expressions that can be used instead of the UCT mathematical expression. We propose two approaches, dubbed as ES partially integrated in MCTS  and ES in MCTS. The former evolves expressions by partially using the MCTS statistical tree to make informed decisions. The latter is fully integrated in the MCTS. We compare these two approaches against 5 variants of the MCTS UCT, 3 variants of the \mbox{*-minimax} family of algorithms and a random controller. We show how the ES in MCTS outperforms these 10 controllers in the Game of Carcassonne. 

The main contribution of this work is to show how EAs can be used in MCTS to evolve in real-time a mathematical expression to be used in the selection policy instead of the well-known UCT. The major advantage of our proposed approach is that it learns how to find a reliable expression in the selection policy without the need to manually tuning it as commonly done in MCTS UCT. There are some interesting works using EAs in MCTS. For example, in~\cite{oxana_2021} we incorporated some notions of MCTS in EAs, an approach commonly referred as rolling horizon.  Cazanave~\cite{Cazenave2007EvolvingMT} used Genetic Programming (GP) for heuristic discovery, whereas Alhejali and Lucas~\cite{alhejali} used GP to enhance MCTS during the rollouts simulations. More aligned to this research is the work by Bravi et al.~\cite{DBLP:conf/evoW/BraviKHT17} where the authors used GP to evolve replacements to the UCB1 formula. They tested their approach using the General Video Game AI framework. They showed the different formulae found by the GP approach. The main limitation in their work was a lack of a more in-depth analysis with other AI techniques. Whereas these and other works including~\cite{baier,lucas} are interesting, ours attempts for the first time carrying out an extensive analysis on the evolution of the UCT by means of evolution strategies in the Game of Carcassonne. Our proposed approach is able to outperform the MCTS UCT along with the rest of the AI controllers used in this study.

The rest of this paper is organised as follows. Section~\ref{sec:background} provides some background in MCTS, EAs and in the game of Carcassonne. Section~\ref{sec:ai:controllers} discusses in detail the controllers used in this work. Section~\ref{sec:experimental} discusses the experimental setup. Section~\ref{sec:results} discusses the results attained by each of the controllers while Section~\ref{sec:discussion} explains the reasons as to why our proposed ES in MCTS outperforms all the controllers. Section~\ref{sec:conclusions} draws some conclusions.

\section{Background}
\label{sec:background}


\subsection{The Mechanics Behind MCTS}

MCTS relies on two key elements: (a) that the true value of an action can be approximated using simulations, and (b) that these values can be used to adjust the policy towards a best-first strategy. The algorithm builds a partial tree, guided by the results of previous exploration of that tree. Thus, the algorithm iteratively builds a tree until a condition is reached or satisfied (e.g., number of simulations, time given to  Monte Carlo simulations), then the search is halted and the best performing action is executed. In the tree, each node represents a state, and directed links to child nodes represent actions leading to subsequent states.
Like many AI techniques, MCTS has several variants. Perhaps, the most accepted steps involved in MCTS are those described in~\cite{6145622}  and are the following: (a) \textit{Selection}: a selection policy is recursively applied to descend through the built tree until an expandable (a node is classified as expandable if it represents a non-terminal state and also, if it has unvisited child nodes) node has been reached, (b) \textit{Expansion}: normally one child is added to expand the tree subject to available actions, (c) \textit{Simulation}: from the new added nodes, a simulation is run to get an outcome (e.g., reward value), and (d) \textit{Backpropagation}: the outcome obtained from the simulation step is backpropagated through the selected nodes to update their corresponding statistics.

Simulations in MCTS start from the root state (e.g., actual position) and are divided into two stages: when the state is added in the tree and when a tree policy is used to select the actions (the selection step is a key element and it is discussed in detail later in this section). A default policy is used to roll out simulations to completion, otherwise. 

One element that contributed to enhance the efficiency in MCTS was the selection mechanism proposed in~\cite{10.1007/11871842_29}. The main idea of the proposed selection mechanism was to design a Monte Carlo search algorithm that had a small probability error if stopped prematurely and that converged to the optimal solution given enough time. That is, a selection mechanism that balances exploration \textit{vs.} exploitation, explained next.

\subsection{Upper Confidence Bounds for Trees}
As indicated previously, MCTS works by approximating “real” values of the actions that may be taken from the current state. This is achieved through building a search or decision tree. The success of MCTS depends heavily on how the tree is built and the selection process plays a fundamental role in this. One particular selection mechanism that has proven to be reliable is the UCB1 tree policy~\cite{10.1007/11871842_29}. Formally, UCB1 is defined as:

\begin{equation}
  UCT = \overline{Q}_j + 2C_E \sqrt{\frac{2 \cdot ln \cdot n}{n_j}}
  \label{eq:uct}
\end{equation}

\noindent where $n$ is the number of times the parent node has been visited, $n_j$ is the number of times child $j$ has been visited and $C_E > 0$ is a constant. In case of a tie for selecting a child node, a random selection is normally used~\cite{10.1007/11871842_29}. The values of $Q_{i,t}$ and thus of $\overline{Q}_j$ are understood to be within [0,1].

Thus, this selection mechanism works due to its emphasis on balancing both exploitation (first term of Eq. 1) and exploration (second term of Eq. 1). That is, every time a node is visited, the denominator of the exploration part increases resulting in decreasing its overall contribution. If, on the other hand, another child node of the same parent node is visited, the numerator increases, so the exploration values of unvisited children increase. The exploration term in Eq. 1 guarantees that each child node has a selection probability greater than zero, which is essential given the random nature of the playouts.

\subsection{Evolutionary Algorithms}

Evolutionary Algorithms (EAs)~\cite{Back:1996:EAT:229867,EibenBook2003}, also known as Evolutionary Computation systems, refer to a set of stochastic optimisation bio-inspired algorithms that use evolutionary principles to build robust adaptive systems. EAs work with a population of $\mu$-\textit{encoded}  potential solutions to a particular problem. Each potential solution, commonly known as an individual, represents a point in the search space, where the optimal solution lies. The population is evolved by means of genetic operators, over a number of generations, to produce better results to the problem.  Each individual is evaluated using a fitness function to determine how good or bad the individual is for the problem at hand. The fitness value assigned to each individual in the population probabilistically determines how successful the individual will be at propagating (part of) its code to future generations. 


The field has its origins in four landmark evolutionary methods. The first author's work in Neuroevolution in Deep Neural Networks~\cite{9383028} provides a nice summary of all of these EAs. In this work we briefly describe the two methods employed in this work:

\subsubsection{Genetic Programming (GP)} This EA was  popularised by Koza~\cite{Koza:1992:GPP:138936}. GP is a form of automated programming where individuals are randomly created by using  functional and terminal sets required to solve a given problem. Multiple types of GP have been proposed in the literature with the typical tree-like structure being the predominant form of GP in EAs.

\subsubsection{Evolution Strategies (ES)} These EAs were introduced in the 1960s by Rechenberg~\cite{Rechenberg10.1007/978-3-642-83814-9_6}. ES are generally applied to real-valued representations of optimisation problems. In ES, mutation is the main operator whereas crossover is the secondary, optional, operator. Historically, there were two basic forms of ES, known as the ($\mu,\lambda$)-ES and the ($\mu+\lambda$)-ES. $\mu$ refers to the size of the parent population, whereas $\lambda$ refers to the number of offspring that are produced in the following generation before selection is applied. In the former ES, the offspring replace the parents whereas in the latter form of ES, selection is applied to both offspring and parents to form the population in the following generation. 

\subsection{The Game of Carcassonne}

Games are great test benchmark problem and are becoming more common in AI to carry out interesting research, ranging from logical-based combinatorial puzzle games~\cite{DBLP:conf/cig/LopezO09,DBLP:conf/gecco/LopezTL07}, board games~\cite{Fred}, to video games~\cite{DBLP:conf/evoW/LopezSOB10,5586508}. In this work, we decide to use a popular board game named Carcassonne. This is a challenging German-style board game that was first released in 2000 and can be played by two to five players.  The objective of the game is to attain a higher score than your opponents. The game consists of 72 playing tiles that are used to build the game board. The square-shaped tiles contain sections of a landscape and can contain several features, such as {farms}, {cities}, {roads} and {monasteries}. Additionally, each player is given a uniform coloured set of game pieces, called {Meeples}. The order of play is decided by the players.

\subsubsection{Placing Tiles}

\begin{figure}[t] 
    \centering\includegraphics[width=0.75\columnwidth]{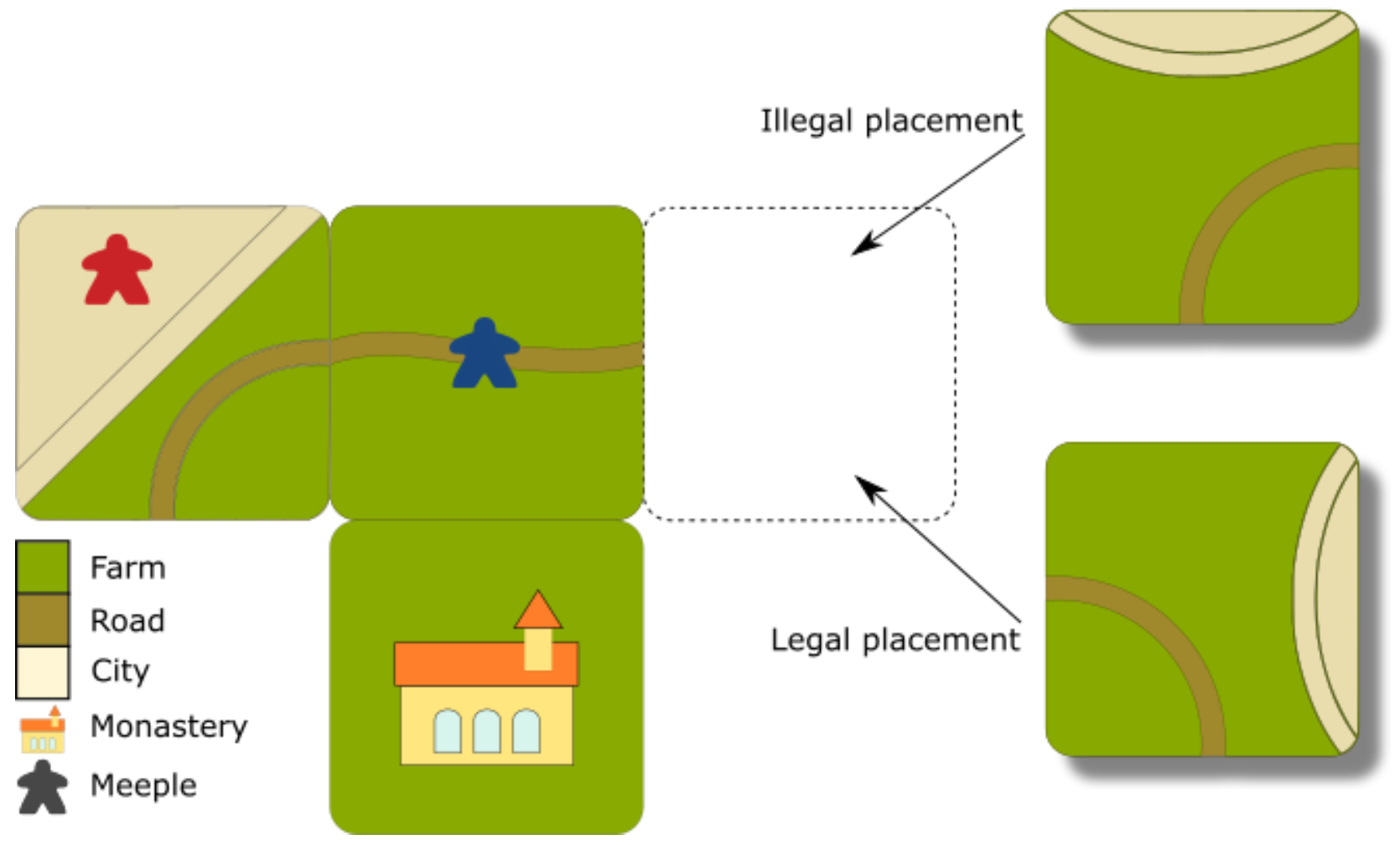}
    \caption{An example of legal and illegal moves when placing a new tile by aligning the sides with the artwork of currently placed tiles.}
    \label{fig:carcassonne_exmaple}
\end{figure}

The game board begins with the same tile being placed face up, and the remaining tiles placed face down in a shuffled deck. In each turn, the current player picks up the tile from the top of the deck and must place the tile adjacent to the any of the previously played tiles on the game board. However, the newly placed tile must extend the features in the adjoining tiles. The tile can be also be rotated as many times, as long as the artwork aligns with all neighbouring tiles. Figure~\ref{fig:carcassonne_exmaple} demonstrates how to legally place tiles.

\subsubsection{Meeples}

The player can then choose to place one of their {meeples} on any of the features in their recently placed tile. The player can only place a {meeple} if they have any left in their inventory. All players begin the game with a total of 7 {meeples} each. Furthermore, the player can choose not to place any {meeple} on their turn, if they wish to do so. However, the player cannot place a {meeple} on a feature that already contains an opponents' {meeple} (or, in fact, one of their own {meeples}). However, it is still possible to have a game feature containing multiple meeples. If two disconnected features of the same type each contain a {meeple}, and are then subsequently connected, both meeples now control the same feature. This can be done as many times throughout the game, and is often employed as a tactic to take control of features from opponents.

\subsubsection{Scoring System}

Scores are added to players' tallies when a feature containing a {meeple} or {meeples} is completed. A {city} is finished when the entire section of outer walls are entirely connected. A {city} tile containing a {pennant} symbol are worth double points. To gain scores from a {road}, both ends of the section must be closed by entering a city, monastery or village. {Monasteries} are only scored once they are entirely surrounded by other tiles. If a completed feature contains multiple sets of {meeples}, the points will be awarded to the players with the most {meeples} within the feature. The same set of {meeples} are then returned to the players.

The game continues until all tiles have been placed down. At the end of the game, {meeples} placed in {farms} gain score for each completed {city} touching the {farm} it is placed in. Further scores are also added for {meeples} placed in incomplete features. The player with the highest score is declared the winner of the game. Table ~\ref{tab:carcassonneScoresTable} provides details on how each feature is scored. A player's score achieved during normal play is known as their {current} score, and their {virtual} score is the player's potential final score from {farms} and incomplete features. Both scores are identical at the end of the game. 

\begin{table}
    \footnotesize
        \centering
        \caption{List of scoring abilities in Carcassonne.}
        \begin{tabular}{|l||p{3.1cm}|p{3.1cm}|}
            \hline
            {Feature}   & {Completion (during play)}        & {End of Game}    \\ \hline \hline
            Farm      & (Not scored during play)        & +3 for each completed city touching field \\ \hline
            Road      & +1 per tile                     & +1 per tile                               \\ \hline
            City      & +2 per tile (+2 per pennant)    & +1 per tile (+2 per pennant)              \\ \hline
            Monastery & +9                              & +1 and +1 for each surrounding tile       \\ \hline
        \end{tabular}
        \label{tab:carcassonneScoresTable}
\end{table}

\section{AI Controllers}
\label{sec:ai:controllers}

\subsection{Monte Carlo Tree Search}

The basic implementation of MCTS is a tree search algorithm that builds a statistical tree with stochastic simulations. It is suitable for any two-player, perfect-information game of finite length. Carcassonne falls into this category. The core idea of MCTS is based on its four primary functions: \textit{Selection}, \textit{Expansion}, \textit{Rollout} and \textit{Backpropagation}. See Section~\ref{sec:background} for details on how these work. The completion of these four stages is known as a single simulation. Increasing the number of simulations will ultimately lead to a more informative search tree. When all simulations conclude, the algorithm will then choose an action associated with a node from the first layer of the tree. Normally the node with the highest action value $Q$ is chosen, which is the approach taken in this work. 


One approach of evaluating the result would be to return a reward value: +1 for a win, -1 for a loss, 0 for a draw. However, both Heyden \cite{Heyden2009IMPLEMENTINGAC} and the first author~\cite{Fred} adopted a more informative result evaluation: the reward value is the difference of scores between the two players.  We compare both approaches. Their results are discussed in Section~\ref{sec:results}.

\subsection{Evolution Strategies Partially Integrated in Monte Carlo Tree Search}

We now turn our attention to the proposed AI controller based on Evolution Strategies (ES) to evolve online the \sloppy{mathematical} expression to be used during the selection phase of the MCTS for the game of Carcassonne. To this end, we use ($\mu$,$\lambda$)-ES, where $\mu=1$ and $\lambda=4$. See Section~\ref{sec:background} for details on how EAs work. We first seed the UCT expression (see Eq.~\ref{eq:uct}) as our initial individual. We then proceed to generate the offspring. Each of these is produced by means of subtree mutation. We evolve a candidate solution in every turn that we need to make a decision and we evolve it for a number of generations. In Section~\ref{sec:experimental} we discuss in detail the parameters used in this work and their corresponding values. 

Our proposed method, as specified before, aims to evolve a mathematical expression that can replace UCT  with the goal to get better or competitive results compared with UCT. Thus, ES is called during the selection step in MCTS. Once a tree node has been selected by our evolved expression, we proceed to compute the fitness of the evolved expression. We do so by performing rollouts as done in MCTS. A key difference is that these values are not replicated from leaves to root and updating the nodes' values in that particular path. We simply keep track of this fitness value. The latter is the score achieved when selecting a node from the MCTS statistical tree and playing the game of Carcassonne. We dubbed this proposed method as Evolution Strategies partially integrated in Monte Carlo Tree Search (ES-p-MCTS, for short).

\subsection{Evolution Strategies in Monte Carlo Tree Search}

We now proceed to describe how the previous  method can be fully integrated in MCTS. We call it Evolution Strategies in Monte Carlo Tree Search (ES-MCTS, for short). Every time we evolve a mathematical expression to potentially be used instead of UCT, we assess it (compute its fitness) by applying rollouts. The value of these rollouts are used to update a \textit{copy} of the MCTS statistical tree, from the selected node to the root including the nodes given in a given branch. That is, this update is performed in the same fashion as done in vanilla MCTS (see Section~\ref{sec:background}). By updating this copy of the statistical tree of the MCTS, we can explore or exploit other parts of the decision space, while at the same time keeping the current status of the game unchanged. We perform ten simulations to compute the fitness of the evolved expression. The fitness of our evolved individual is the average of these ten simulations. We then use the best (evolved) individual in the MCTS algorithm instead of the UCT expression.

\subsection{Minimax}

The classic \textit{minimax} search is expanded for stochastic games as the \textit{expectimax}  algorithm \cite{ballard1983minimax,hauk2004m}. Expectimax handles chance nodes by weighting their minimax values according to the probabilities of the respective events. The *-minimax family of algorithms, including  Star1, Star2, and Star2.5, are expectimax variants that use an alpha-beta pruning technique adapted for stochastic trees.


 In the Star1 algorithm, the theoretical maximum value $U$ and the theoretical minimum value $L$ are used as the guess for the worst and best scenarios of the chance nodes that have not been evaluated in an attempt to prune the tree if the predicted values fall outside an $\alpha\beta$ window as in alpha-beta pruning. In the worst-case scenario, no nodes are pruned and the search behaves as the normal expectimax.

 Star2 is meant for \textit{regular *-minimax games}, in which the actions for each player are influenced by a stochastic event at the beginning of each turn. Examples of regular *-minimax games are Backgammon, Catan, Monopoly, and Carcassonne. In Star2, the first node is evaluated and used as the guessed value for the rest of the sister nodes to prune as in Star1, in the {probing phase}. Thus, ordering of the actions is required to get more reliable results and to prune more often, leading to a faster computational calculation. The actions available from each state are ordered as soon as each state is reached for the first time according to how promising they are (best to worst). The ordering is done following a heuristic that is cheaper than the simulation of the action and the evaluation of the resulting state. If the probing phase fails to achieve a cut-off, the search behaves as the Star1.

 A \textit{probing factor} $f>1$ can be predefined in the Star2.5 algorithm. The probing factor determines the number of nodes to be evaluated during the probing phase. In other words, a $f=0$ stands for Star1, $f=1$ refers to Star2 and $f>1$ is the Star2.5. We used all these three variants to see which one yields the best results. 

 \subsection{Random}

We also use a controller that chooses moves at random. This controller chooses an action from a set of available possible moves with uniform probability during its turn. This player will predominantly be used as a baseline to demonstrate that the remaining AI controllers are well-informed players that perform intelligent moves

\section{Experimental Setup}
\label{sec:experimental}

\subsection{Function and Terminal Sets}
We are interested in evolving mathematical expressions that can be used instead of the UCT equation, shown in Eq.~\ref{eq:uct}. To this end, we define the function set and the terminal set as follows. $F = \{+,-,\times,\div,\log,\sqrt { }\}$ and $T =  \{Q(s,a),N(s),N(s,a),C_E\}$, where $N(s)$ is the number of visits to the node from the MCTS search tree, $N(s,a)$ is the number of visits to a child node, $Q(s,a)$ is the child's node action-value and $C_E$ is the exploration-exploitation constant. When $C_E$ is chosen to be mutated, it can take a random value from the following set $r = \{0.25, 0.5, 1, 2, 3, 5, 7, 10\}$. The division operator is protected against division by zero and will return 1 for any divisor less than 0.001. Similarly, the natural log and square root operators are protected by taking the absolute values of input values.

\begin{table}[tb]
\centering
\caption{Summary of Parameters used in our experiments.}
\resizebox{0.90\columnwidth}{!}{ 
\small\begin{tabular}{|l|r|} \hline 
\emph{Parameter} & \emph{Value} \\ \hline \hline

\multicolumn{2}{|c|}{*-Minimax}\\ \hline
Max Depth & 2 for Star 1, 2 and 2.5 \\ \hline
Lower Bound & $L=-100$, for Star 1, 2 and 2.5 \\ \hline
Upper Bound & $U=100$, for Star 1, 2 and 2.5 \\ \hline
Probing factor & 0, 1, 4, for Star 1, 2 and 2.5, respectively \\ \hline

\multicolumn{2}{|c|}{ES-p-MCTS and ES-MCTS}\\ \hline
($\mu$,$\lambda$)-ES &  $\mu=1$, $\lambda=4$  \\ \hline
Generations & 20 \\ \hline
Type of Mutation & Subtree (internal node), Point (leaf) \\ \hline
Mutation Rate & One per individual \\ \hline
Initialisation Method & Seeded and mutated  \\ \hline

\multicolumn{2}{|c|}{MCTS}\\ \hline
No. of simulations & 400 \\ \hline
$C_E$ & $\{0.25,0.5,1,2,3\}$ \\ \hline


\end{tabular}
}
\label{tab:parameters}
\end{table}

\subsection{League Competition Scoring System}

The League competition scoring system is inspired by the system adopted by most major professional sport leagues, where points are awarded depending on the result of a game between two player (in our case,  between two controllers). The points awarded to a game of Carcassonne based on a win, a loss, a draw  or bonus win (or loss)  points are shown in Table~\ref{table:points}. A bonus point is attained for game win ratios greater than 75:25 and for loss ratios greater than 45:55. Furthermore, this system also considers each controller being first or second player. The latter leads to a fair comparison by letting a controller play as first player a number of games and then act as second player for the same number of games.

\subsection{Extensive Empirical Experimentation}

To  obtain  meaningful  results,  we  carried  out  an  extensive empirical experimentation: 1,100 independent games. This was done as follows. We first group the MCTS controllers to play against each other.  Given that we have five variants for the MCTS, we performed 500 independent games (20 matches of 25 games each). We also grouped the three Star algorithms. This led to have 150 independent games (6 matches of 25 games each).

Finally, we compared our two proposed ES-based MCTS controllers against the best controllers of MCTS and Star algorithms. We also considered a random controller and the MCTS controller with the highest average point difference value. This led to have 450 independent games (30 matches of 15 games each). At every turn of each of these 1,100 games, we performed 400 simulations. We executed our games using a desktop computer with Intel Core i5 CPU clocked at 3.40GHz and 8GB RAM. We completed running these experiments over a period of four weeks. The results of this extensive empirical experimentation is discussed in the next paragraphs.

\begin{table}[tb]
  \begin{center}
\caption{Points awarded to a game of Carcassonne based on a win, a loss or a draw.}
\begin{tabular}{|l|l|r|} \hline
  Acronym     &  Description  & Points\\ \hline \hline

BWP     & Number of Bonus Win Points                                                                        & 1      \\ \hline
BLP     & Number of Bonus Loss Points                                                                       & 1      \\  \hline
W       & Number of wins                                                                                    & 4      \\   \hline
L       & Number of losses                                                                                  & 0      \\ \hline
D       & Number of draws                                                                                   & 2       \\  \hline
PD      & \begin{tabular}[c]{@{}l@{}}Average Point Difference\\  PD=Player Score-Opponent Score\end{tabular} &        \\  \hline
Points  & Points tally of Controller                                                                        &        \\  \hline
\end{tabular}
\label{table:points}
\end{center}
\end{table}

\section{Discussion of Results}
\label{sec:results}

\begin{table}[tb]
  \begin{center}
    \caption{Number of Wins (W), Losses (L) and Draws (D) by using two different reward systems. Top: using +1, 0 and -1 as reward values for W, D and L, respectively. Bottom: using difference of scores as reward values between the two players. No. of simulations: 400. No. of independent games: 50.}
    \resizebox{\columnwidth}{!}{%
        \begin{tabular}{|c||r|r|r|r|r|r|r|}
        \hline
        \multirow{2}{*}{Controller  }
            & \multirow{2}{*}{Reward system}     
            & \multicolumn{3}{c|}{Player 1}  &   \multicolumn{3}{c|}{Player 2}      \\
            
        &   & W & L & D & W & L & D  \\    
        \hline \hline
        C1          & +1,0,-1   & 2     & 48    & 0     & 4     & 45    & 1                           \\ \hline
        C2          & Diff. of scores       & 45    & 4     & 1     & 48    & 2     & 0                             \\ \hline
        \end{tabular}
        }
    \label{table:mcts:reward}
    \end{center}

\end{table}

\subsection{Reward Systems}

When MCTS is used in a two-player game as done in this work, the reward value is normally defined as +1, 0 and 1 when the player wins, draws or losses against its opponent, respectively. However, as indicated in Section~\ref{sec:ai:controllers}, a more informative case seems to be more beneficial. Specifically, when the difference of scores between the two players is used as reward value.

Table~\ref{table:mcts:reward} shows the results when using these reward systems: +1, 0, -1 denoted by Controller 1 (C1) and  the difference of scores between the two players denoted by Controller 2 (C2). We used these controllers to compete against each other for 400 simulations using 50 independent games. When C1 is used as first player, it only wins 2 games against C2 that is the second player. In the latter case, C2 wins 48 games (see bottom right-hand side of the table). When C2 is used as a first player and C1 as a second player, we can see that C2 wins 45 games and C1 only wins 4 games (see top right-hand side of the table). In summary, C2 (difference of scores between the two players to be used as a reward value) yields the best results and it is the reward system used in the rest of the experiments.

\subsection{Monte Carlo Tree Search and Different Values for UCT}

The performance of MCTS is greatly determined by the \textit{exploitation-exploration} constant $C_E$ in the UCT function (see Eq.~\ref{eq:uct}). Increasing this value will result in less visited nodes or nodes with smaller action-values to be explored more. A lower value for $C_E$ will closely resemble a greedy algorithm that will exploit the best possible action during each simulation. A round-robin tournament with five MCTS controllers with different $C_E$ values (see Table~\ref{tab:parameters} for the values used in this experiment) is adopted in this work to determine what value yields the most competitive MCTS' results.

Table~\ref{table:mcts:matches} shows the results of the matches between pairs of MCTS controllers using different values for $C_E$ in the UCT as normally adopted in MCTS. When $C_E=2$ and playing as Player 1 (left-hand side of the table), we can see that this controller won all the matches against the rest of the controllers. More specifically it won 16, 19, 13 and 16 games out of the 25 games that each match consists of. When this controller plays as a second player (top of the table), it wins only one match, which is against MCTS $C_E=3$, where it wins 13 of the 25 games. Notice that in some cases the sum of the wins shown in the cells is not equal to 25 games. This denotes a draw and it is computed as the difference of the sum of the wins and the the total number of games (25) in each match. Continuing our attention on this controller (MCTS $C_E=2$), we can see that it losses three matches when playing as a second player against MCTS $C_E=\{0.25,0.5,1\}$. Furthermore, there is not a draw in any of these matches (either MCTS $C_E=2$ being first or second player).

We then use this information to rank each player. This is shown in Table~\ref{table:mcts:results}. Two elements determine this ranking: points attained by a controller (see Table~\ref{table:points}) and in case of a tie, we use  the average point difference between a controller and the rest of the controllers (denoted as PD in Table~\ref{table:mcts:results}). As can be seen from the table, when MCTS uses $C_E=2$ in its UCT formula, the controller wins five matches, losses three and there are no draws. This gives 21 points (see first row of Table~\ref{table:mcts:results}). This is easily computed by considering Table~\ref{table:points}: 4 points for a win (4$\times$5 wins) and 1 point for a win with ratio greater than 75:25. The same number of points is attained when $C_E=0.5$. This is the result of winning 5 matches (4 points $\times$ 5) and 1 point for a loss within a loss ratio of 45:55. The tie breaker is determined then by the average point difference, denoted by PD in the table, where $C_E=\{2,0.5\}$ achieve 12.48 and 8.2, respectively. 

We can observe from Tables~\ref{table:mcts:matches} and~\ref{table:mcts:results} that the best performing MCTS controller is when $C_E=2$ is defined in the UCT formula. This is followed closely when $C_E=0.5$. The bottom three controllers are when $C_E=\{1,0.25,3\}$, from best to worst. However, it is interesting to note that when $C_E=1$, this attains the highest PD value among all the five controllers.

\begin{table}[htbp]
    \caption{Number of wins out of 25 games, for each of the 20 pair matches (25*20 = 500 independent games) when using \textbf{MCTS} and five different $C_E$ values.}
    \begin{center}
    \resizebox{1.0\columnwidth}{!}{%
        \begin{tabular}{ll||c|c|c|c|c|}
        \cline{3-7}
                                &   & \multicolumn{5}{c|}{{Player 2}}                                       \\ \cline{3-7} 
        \multicolumn{1}{l}{}    &   & $C_E=0.25$ & $C_E=0.5$ & $C_E=1$ & $C_E=2$  & $C_E=3$ \\ \hline \hline
        
        \multicolumn{1}{|l|}{\multirow{5}{*}{\rotatebox[origin=c]{90}{{Player 1}}}} 
        
                                        &  $C_E=0.25$ & -       & 10-14  & 12-12  & 15-10   & 12-13       \\ \cline{2-7} 
        \multicolumn{1}{|l|}{}          &  $C_E=0.5$  & 11-13   & -      & 15-10  & 15-10   & 15-9       \\ \cline{2-7} 
        \multicolumn{1}{|l|}{}          &  $C_E=1$    & 9-16    & 11-14  & -      & 15-10   & 17-8        \\ \cline{2-7} 
        \multicolumn{1}{|l|}{}          &  $C_E=2$    & 16-9    & 19-6   & 13-11  & -       & 16-8        \\ \cline{2-7} 
        \multicolumn{1}{|l|}{}          &  $C_E=3$    & 17-8    & 14-11  & 10-15  & 12-13   & -           \\ \hline 
        \end{tabular}
    }
    \label{table:mcts:matches}
    \end{center}

        \caption{\textbf{Ranking of the MCTS} controllers based on Points as determined by Wins (W), Losses (L), Draws (D), Bonus Win Points (BWP) and Bonus Loss Points (BWP). See Table~\ref{table:mcts:matches} for full details of results per match.}

    \begin{center}
    \resizebox{\columnwidth}{!}{%
        \begin{tabular}{|c||l|r|r|r|r|r|r|r|}
        \hline
        {Pos} & {MCTS}  & {Points} & {BWP} & {BLP} & {W} & {L} & {D} & {PD}     \\ \hline \hline
        1   & $C_E=2$                     & 21     & 1   & 0   & 5 & 3 & 0 & +12.48       \\ \hline
        2   & $C_E=0.5$                   & 21     & 0   & 1   & 5 & 3 & 0 & +8.2         \\ \hline
        3   & $C_E=1$                     & 15     & 0   & 1   & 3 & 4 & 1 & +20.28       \\ \hline
        4   & $C_E=0.25$                  & 15     & 0   & 1   & 3 & 4 & 1 & -19.96       \\ \hline
        5   & $C_E=3$                     & 13     & 0   & 1   & 3 & 5 & 0 & -21          \\ \hline
        \end{tabular}
    }    
    \label{table:mcts:results}
    \end{center}
\end{table}

\subsection{Performance of Star1, Star2 and Star2.5}
\begin{table}[tb]
  \caption{Number of wins out of 25 games, for each of the 6 pair matches (25*6 = 150 independent games) when using \textbf{Star1, Star2 and Star2.5}.}
  
    \begin{center}
        \begin{tabular}{ll||c|c|c|}
        \cline{3-5}
                                &   & \multicolumn{3}{c|}{{Player 2}}                                       \\ \cline{3-5} 
        \multicolumn{1}{l}{}    &   & Star1 &  Star2 &  Star2.5  \\ \hline \hline
        
        \multicolumn{1}{|l|}{\multirow{3}{*}{\rotatebox[origin=c]{90}{{\scriptsize{Player 1}}}}} 
        
                                        &  Star1    & -        & 14-11  & 12-12            \\ \cline{2-5} 
        \multicolumn{1}{|l|}{}          &  Star2    & 11-14    & -      & 14-10            \\ \cline{2-5} 
        \multicolumn{1}{|l|}{}          &  Star2.5  & 11-14    & 9-16   & -                \\ \hline
        \end{tabular}
    \label{table:minimax:matches}
    \end{center}
    \caption{\textbf{Ranking of the Star} controllers based on Points as determined by Wins (W), Losses (L), Draws (D), Bonus Win Points (BWP) and Bonus Loss Points (BWP). See Table~\ref{table:minimax:matches} for full details of results per match.}

    \begin{center}
    \resizebox{\columnwidth}{!}{%
        \begin{tabular}{|c||l|r|r|r|r|r|r|r|}
        \hline
        {Pos} & {Player}  & {Points} & {BWP} & {BLP} & {W} & {L} & {D} & {PD}     \\ \hline \hline
        1   & Star1          & 14    & 0   & 0   & 3 & 0 & 1 & +8        \\ \hline
        2   & Star2         & 8     & 0   & 0   & 2 & 2 & 0 & -1.16     \\ \hline
        3   & Star2.5        & 2     & 0   & 0   & 0 & 3 & 1 & -6.84     \\ \hline
        \end{tabular}
    }    
    \label{table:minimax:results}
    \end{center}
\end{table}

Let us now focus our attention on the performance achieved by the Star controllers. With only three controllers and using a robin-round tournament match, as done with all the experiments reported in this section, we have 6 matches (one match is composed of 25 games). The results of these matches are shown in Table~\ref{table:minimax:matches}. From this table, we can see that Star1, acting as first player (left-hand side of the table) is able to win a match against Star2 and it attains a draw against Star2.5.  When Star1 is now Player 2, shown in the top of the table, we can observe that this controller wins all the matches (11-14, in both cases, against Star2 and Star2.5).

This and the rest of the summary of wins, losses and draws, among other informative values including number of bonus wins/loss points (referred as BWP/BLP), and average point difference (PD) between controllers, are shown in Table~\ref{table:minimax:results}.  Star1 yields the best results among all the controllers, without losing any match. This is followed by Star2, with 2 wins and 2 losses. Finally, in last place, we have Star2.5 with no win matches.

\subsection{Comparison of performance of Evolution Strategies-based MCTS controllers against the rest of the controllers}

\begin{table*}

        \caption{Number of wins out of 15 games, for each of the 30 pair matches (15*30 = 450 independent games) when using \textbf{\SECONDEA, \FIRSTEA, MCTS $(C_E=\{1,2\})$,  Star1 and Random}.}

  \begin{center}
    \resizebox{0.65\textwidth}{!}{
               \begin{tabular}{ll||c|c|c|c|c|c|}
         \cline{3-8}   &    & \multicolumn{6}{c|}{{Player 2}}                              \\ 
\cline{3-8}  \multicolumn{1}{l}{}  &  & ES-MCTS & ES-p-MCTS & MCTS ($C_E=1$) & MCTS ($C_E=2$)   & Star1    & Random    \\ \hline 

\cline{2-8} 
\multicolumn{1}{|l|}{\multirow{6}{*}{\rotatebox[origin=c]{90}{{Player 1}}}}  &   ES-MCTS         & -     & 15-0  & 12-3  & 12-3   & 9-6   & 15-0  \\
\cline{2-8} 
\multicolumn{1}{|l|}{}  &   ES-p-MCTS          & 2-13  & -     & 4-11  & 2-13   & 1-14  & 15-0  \\
\cline{2-8} 
\multicolumn{1}{|l|}{}  &   MCTS ($C_E=1$)          & 6-9   & 15-0  & -     & 9-6    & 7-8   & 15-0  \\
\cline{2-8} 
\multicolumn{1}{|l|}{}  &   MCTS ($C_E=2$)         & 4-11  & 15-0  & 7-8   & -      & 6-9   & 15-0  \\
\cline{2-8}       
\multicolumn{1}{|l|}{}  &   Star1           & 6-7   & 13-2  & 10-5  & 5-10    & -     & 15-0  \\
\cline{2-8} 
        \multicolumn{1}{|l|}{}  &   Random      & 0-15  & 0-15  & 0-15  & 0-15   & 0-15  & -     \\ \hline 
        \end{tabular}
    }
    \label{table:all:matches}
  \end{center}

\end{table*}

We now turn our attention on the performance achieved by the Evolutionary Strategies variants in MCTS. We compare each of these by playing matches against each other as well as playing matches against those controllers discussed previously that yield the best results in their respective groups. That is, we use the best controller from MCTS ($C_E=2$) and Star algorithms (Star1). We also include a random controller. Because we are primarily interested to know whether our proposed ES-MCTS variants yield competitive or better results against MCTS or not, we further include MCTS with $C_E=1$: this yields the highest value for the average point difference (PD) between two controllers (see Table~\ref{table:mcts:results}) and was the third best controller in its MCTS group.  All these controller led to have ten matches between a particular controller against the rest of the controllers, totaling 30 matches of 15 games each, leading to run 450 independent games of 400 simulations at each turn. The results of these matches are shown in Table~\ref{table:all:matches}.

Let us  focus our attention in the Evolution Strategy in MCTS, denoted as ES-MCTS in Table~\ref{table:all:matches}, we can observe that when this controller plays as second player, it beats the rest of the controllers. When ES-MCTS goes as first player, we can see that it is able to beat all the controllers too. Thus, the total number of wins for all ten matches is ten, as specified under the $W$ column in Table~\ref{table:all:results}. Moreover, there were nine matches with a 75:25 ratio leading to nine points. Thus, the total number of points attained by ES-MCTS is 49 (10 win matches $\times$ 4 points, plus 9 bonus win points). The MCTS variants with $C_E=\{1,2\}$ attain six and five win matches, respectively. When adding their bonus points (either by wins or losses), we can see that MCTS with $C_E=1$ attains 29 points and MCTS with $C_E=2$ achieves 26 points. These are ranked 3$^{rd}$ and 4$^{th}$, respectively, as can be seen in Table~\ref{table:all:results}. Finally, ES-p-MCTS and Random show poor performance with ten and zero points, respectively. 

\subsection{Simulations carried out when using ES-MCTS and ES-p-MCTS}

As discussed in Section~\ref{sec:ai:controllers}, the ES-based MCTS variants evolves potential solutions over a small number of generations. This, however, does not mean that these variants have an advantage over the other AI controllers used in this study. Specifically, when comparing MCTS-ES against MCTS, ranked 1$^{st}$ and 3$^{rd}$, respectively, when considering the best AI controllers (see Table~\ref{table:all:results}). Whereas it is true that MCTS-ES performs more simulations, these are not considered in the statistical tree built by MCTS. Rather MCTS-ES uses a \textit{copy} of this statistical tree. Thus, when the EA simulations finish, the MCT-ES statistical tree is the result of MCTS simulations only, which is comparable to the MCTS UCT.

\subsection{Statistically Significant Results}

We carried out a statistical analysis on the results attained at each group. Specifically, the statistical significance for the results on the scores attained at each game for each group summarised in Tables~\ref{table:mcts:results},~\ref{table:minimax:results} and~\ref{table:all:results} was computed using Wilcoxon Test at 95\% level of significance, independently comparing the best MCTS controller (MCTS $C_E=2$ as seen in Table~\ref{table:mcts:results}), the best Star algorithm (Star1 as seen in Table~\ref{table:minimax:results}) and the best controller from Table~\ref{table:all:results} (ES-MCTS), in each case against each of the controllers in their groups (in their corresponding tables). For MCTS $C_E=2$, this is statistically significant with those results attained by MCTS $C_E=\{1,0.25,3\}$ and not statistically significant with MCTS $C_E=0.5$. For Star1, this  is statistically significant with those results attained by Star2 and Star2.5. Finally,  ES-MCTS  is statistically significant with those results attained by all the controllers in their group (these are listed in Table~\ref{table:all:results}).

\begin{table}[tb]
    \caption{\textbf{Ranking of the  \SECONDEA, \FIRSTEA, Star1, MCTS ($C_E=\{1,2\}$) and Random} controllers based on Points as determined by Wins (W), Losses (L), Draws (D), Bonus Win Points (BWP) and Bonus Loss Points (BWP). See Table~\ref{table:all:matches} for full details of results per match.}

    \begin{center}
      \resizebox{\columnwidth}{!}{%

        \begin{tabular}{|c||l|r|r|r|r|r|r|r|}
        \hline
        {Pos} & {Player}  & {Points} & {BWP} & {BLP} & {W} & {L} & {D} & {PD}     \\ \hline \hline
        
        1   & ES-MCTS               & 49     & 9    & 0   & 10 & 0  & 0 & +381.67     \\ \hline
        2   & Star1                 &   34   & 4    & 2   & 7  & 3  & 0 & +213.66     \\ \hline
        3   & MCTS ($C_E=1$)          &   29   & 4    & 1   & 6 & 4  & 0 & +218.47     \\ \hline
        
        4   & MCTS ($C_E=2$)          &   26   & 5    & 1   & 5  & 5  & 0 & +203.60     \\ \hline
       
        5   & ES-p-MCTS             &   10   & 2    & 0   & 2  & 8 & 0 & -326.06     \\ \hline
        6   & Random                & 0      & 0    & 0   & 0  & 10 & 0 & -1186.87    \\ \hline
        \end{tabular}
    }    
    \label{table:all:results}
    \end{center}

  \caption{Five most common expressions and occurrences that these were yielded by the ES partially integrated in MCTS (ES-p-MCTS) and ES in MCTS (ES-MCTS) controllers.}
    \begin{center}
    \resizebox{0.95\columnwidth}{!}{%
        \begin{tabular}{|l||l|r||l|r|}
        \hline
        {Num} & {ES-p-MCTS} & {Count} &  {ES-MCTS} & {Count}  \\ \hline \hline
        1  & $\overline{Q}_j + 2C_E \sqrt{\frac{2 \cdot \log n}{n_j}}$ & 263 & $\overline{Q}_j + 2C_E \sqrt{\frac{2 \cdot \log n}{n_j}}$    & 92 \\ \hline
        2  & $\sqrt{n}$                                 & 28  & $\sqrt{\overline{Q}_j}$                  & 27     \\ \hline
        3  & $\sqrt{2}$                                 & 24  & $\log \overline{Q}_j$                    & 24       \\ \hline
        4  & $\sqrt{\overline{Q}_j}$                                 & 19  & $\log n_j$                    & 21       \\ \hline
        5  & $\log n_j$                                   & 16  & $\overline{Q}_j + \sqrt{\overline{Q}_j}$              & 19      \\ \hline \hline
        
        {Total}   &   \multicolumn{2}{r||}{8517}   &   \multicolumn{2}{r|}{8517}  \\ \hline 
        {Unique}  &   \multicolumn{2}{r||}{7444}   &   \multicolumn{2}{r|}{7607} \\ \hline
        \end{tabular}
    }    
    \label{table:evolved:expressions}
    \end{center}
    
\end{table}

\section{Why does ES-MCTS outperform ES-p-MCTS?}
\label{sec:discussion}


We have already seen from Tables~\ref{table:all:matches} and~\ref{table:all:results}, how the proposed Evolution Strategies  in MCTS (ES-MCTS) outperformed the rest of the controllers. One of the reason why this occurs is due to the ability of the ES-MCTS approach to use the statistical tree of MCTS along with the ability to create mathematical expressions that can be used instead of the UCT expression (see Eq.~\ref{eq:uct}), as  normally adopted in MCTS.

To demonstrate the latter, we kept track of all the independent games among all the controllers used in Tables~\ref{table:all:matches} and~\ref{table:all:results}. In particular, we recorded those evolved expressions generated by our ES variants. In Table~\ref{table:evolved:expressions}, we show the five most common evolved expressions yielded by ES partially integrated in MCTS (ES-p-MCTS) and ES in MCTS (ES-MCTS). The table also shows the number of occurrences of these five most common evolved expressions as well as the total number of expressions evolved by our controllers and the number of unique expressions found by each of these ES variants. These two are shown at the bottom of the table.

The unchanged UCT formula is the most popular with each of the controllers. We can see that the ES partially integrated in MCTS (ES-p-MCTS) uses the UCT formula more often (263 times) compared to ES-MCTS (92 times). This seems to indicate that ES-p-MCTS fails at producing a competitive or better formula compared to UCT. When we take a closer inspection to the rest of the most common evolved expressions produced by ES-p-MCTS, we can see that only one of these make use of the action-value $\overline{Q}_j$. This is a key element to make intelligent decisions. 

When we take a look at the expressions generated by the ES-MCTS shown in the right of Table~\ref{table:evolved:expressions}, we can see that this situation changes radically. For instance, in all five common expressions, but one, evolved by the ES in MTS (ES-MCTS) the use of the value-action  $\overline{Q}_j$ is evident.  The total number of generated expressions were the same for both ES-based controllers. 

\section{Conclusions and Future Work}
\label{sec:conclusions}

Monte Carlo Tree Search (MCTS) is a sampling best-first method to search for optimal decisions. A breakthrough in MCTS was the adoption of the UCT, that yields extraordinary results provided that this is well calibrated and enough simulations are employed in MCTS. We have proposed (i) Evolution Strategies (ES) partially integrated in MCTS (ES-p-MCTS) and (ii) ES in MCTS (ES-MCTS). The latter uses a \textit{copy} of the MCTS statistical tree whereas the former does not replicate the reward value attained by the simulations, as normally adopted in MCTS. We have seen how the proposed ES-MCTS is able to outperform all the ten controllers used in this work, including robust MCTS controllers. 

We believe that the information retrieved from rollouts is underutilised in MCTS. From preliminary research, it is clear that we can use this taking inspiration from GP semantic-based approaches~\cite{DBLP:conf/gecco/GalvanS19,DBLP:journals/corr/abs-2012-04717,9308386,DBLP:conf/ppsn/LopezMES16,GALVAN2021108143,snocec2021,DBLP:journals/gpem/UyHOML11}. These results show its adoption in MCTS can have a positive impact in the results on the Game of Carcassonne.

\section*{Acknowledgments}
The authors thank the Department of Computer Science at Maynooth University for financial support. The authors thank Fred Valdez Ameneyro for providing the implementation of the Game of Carcassonne. The authors also thank the reviewers for their useful comments on the paper.

\bibliographystyle{abbrv}
\bibliography{2021_SSCI_EdgarGalvan}

\end{document}